\documentclass[a4paper,fleqn]{cas-dc}

\usepackage[numbers]{natbib}
\usepackage{algorithmic}
\usepackage{amsmath,amssymb,amsfonts}
\usepackage{textcomp}
\usepackage{acro}
\usepackage{color}
\usepackage{verbatim}
\usepackage{url}
\usepackage{xcolor}
\usepackage{multirow}
\usepackage{color, colortbl}
\usepackage{textcomp}
\usepackage{hyperref}
\usepackage{subfig}

\newcommand{\mj}[1]{{\color{orange}#1}}

\DeclareAcronym{PLC}{
    short=PLC,
    short-plural-form=PLCs,
    long=Programmable Logic Controller,
    long-plural-form=Programmable Logic Controllers,
    short-indefinite=a,
    long-indefinite=a
}

\DeclareAcronym{IWTS}{
  short = IWTS,
  long  = Industrial Water Treatment Systems
}

\DeclareAcronym{DL}{
  short = DL,
  long  = Deep Learning
}

\DeclareAcronym{LSTM}{
  short = LSTM,
  long  = Long Short-Term Memory
}

\DeclareAcronym{Ti-iLSTM}{
  short = Ti-iLSTM,
  long  = TinyDL-based incremental LSTM
}

\DeclareAcronym{IIoT}{
  short = IIoT,
  long  = Industrial Internet of Things
}

\DeclareAcronym{TinyDL}{
  short = TinyDL,
  long  = Tiny Deep Learning
}

\DeclareAcronym{RSS}{
  short = RSS,
  long  = Resident Set Size
}

\DeclareAcronym{ML}{
  short = ML,
  long  = Machine Learning
}

\DeclareAcronym{VIF}{
  short = VIF,
  long  = Variance Inflation Factor
}

\def\tsc#1{\csdef{#1}{\textsc{\lowercase{#1}}\xspace}}
\tsc{WGM}
\tsc{QE}
\tsc{EP}
\tsc{PMS}
\tsc{BEC}
\tsc{DE}

\begin{document}
\let\WriteBookmarks\relax
\def\floatpagepagefraction{1}
\def\textpagefraction{.001}
\shorttitle{Ti-iLSTM: A TinyDL Approach for Logic-Level Anomaly Detection in Industrial Water Treatment Systems}
\shortauthors{M. Joshi et~al.}

\title [mode = title]{Ti-iLSTM: A TinyDL Approach for Logic-Level Anomaly Detection in Industrial Water Treatment Systems}                      



\author[1]{Mandar Joshi} [type=author,auid=000,bioid=1,
                        orcid=0009-0005-0009-7310]
\ead{mj636@students.waikato.ac.nz}


\affiliation[1]{organization={University of Waikato},
                city={Hamilton},
                country={New Zealand}}

\author[1]{Farzana Zahid}[type=author,auid=000,bioid=1,
                        orcid=0000-0002-9277-0787]
                        \cormark[1]
\ead{farzana.zahid@waikato.ac.nz}

\author[1]{Judy Bowen} [type=author,auid=000,bioid=1,
                        orcid=0000-0003-2815-8267]
\ead{judy.bowen@waikato.ac.nz}


\affiliation[2]{organization={Auckland University of Technology},
                city={Auckland},
                country={New Zealand}}

\author[2]{Matthew M.Y. Kuo}[type=author,auid=000,bioid=1,
                        orcid=0000-0001-7269-5874]
\ead{matthew.kuo@aut.ac.nz}

\affiliation[3]{organization={Aalto University},
                city={Espoo},
                country={Finland}}
\affiliation[4]{organization={Luleå University of Technology},
                city={Luleå},
                country={Sweden}}

\author[3,4]{Valeriy Vyatkin}[type=author,auid=000,bioid=1,
                        orcid=0000-0002-9315-9920]
\ead{vyatkin@ieee.org}

\author[4]{Emil Karlsson}[type=author,auid=000,bioid=1,
                        orcid=0009-0007-7445-926X]
\ead{Emil.Karlsson@aalto.fi}

\cortext[cor1]{Corresponding author}


\begin{abstract}
\leavevmode \ac{IWTS} are safety critical cyber-physical infrastructures and due to increased connectivity, these systems are exposed to cyber threats that can manipulate process behaviour without creating obvious devices' outliers. In particular, logic-layer deception anomalies can preserve numerically plausible measurements while breaking expected cause-and-effect relationships in the control process. These attacks are difficult to detect using threshold-based monitoring or require heavy server-oriented anomaly detection models. This paper explores the potential of \ac{TinyDL} to provide lightweight on-device logic-level anomaly detection for resource constrained \acp{PLC}. We propose a novel framework, \ac{Ti-iLSTM} which optimises the memory and space foot print of \ac{LSTM}, to detect logic-layer inconsistencies in \ac{PLC} based \ac{IWTS}. Experiments on the publicly available SWaT dataset show that the optimised model achieves high detection performance (F1-score~$\approx 0.983$ and ROC-AUC~$\approx 0.998$). A deployment-style validation on the WADI dataset confirms that the proposed light-weight framework remains applicable beyond a single dataset. The research demonstrates that combining logic-aware supervision with \ac{TinyDL} sequence learning creates an efficient and accurate anomaly detection suitable for resource constrained \acp{PLC} in industrial environments.
\end{abstract}



\maketitle
\acresetall 
\section{Introduction}
\label{sec:intro}
Industrial systems are rapidly transitioning to automated, heterogeneous and interconnected environments due to the integration of digital control technologies and networked communication infrastructure  \cite{abraham2025towards, murray2017convergence}. Such systems
serve as the backbone of critical infrastructure, supporting essential services such as water
supply, energy distribution, transportation services, and manufacturing. Modern industrial systems, comprising 
thousands of industrial sensors, actuators, \acp{PLC}, and embedded devices communicating in real-time, have given rise to the \acf{IIoT}. This shift has enabled unprecedented levels of automation and operational efficiency. 
However, as \ac{IIoT} systems provide vital services and operate continuously, ensuring their security and reliability is critical to operational integrity \cite{zahid2025light,gulzar2024analytical, tuptuk2021systematic,abraham2025towards}.

Among various \ac{IIoT}-based critical infrastructures, \acf{IWTS} represent one of the most
safety-sensitive/\ac{IIoT} environments where operations depend heavily on digital  monitoring and controlling of  physical
processes \cite{tuptuk2021systematic}. Water treatment plants process raw water
through filtration and chemical processing before storing the treated product
for distribution purposes \cite{mathur2016swat,aias20240006}. \ac{IWTS} consist of sensors to determine water levels and flow rates,
actuators including pumps and valves that control water movement, and \ac{PLC}s that form the core computational component of \ac{IWTS} \cite{kang2016model,Whatisap37:online}. 
During normal operation, sensors, actuators, and \ac{PLC} programs
interact according to fixed and repeatable process rules, resulting in
structured patterns in the recorded sensor and actuator
data \cite{mathur2016swat}. Additionally, \ac{IWTS} exhibit strong temporal behaviour, where actions to change the states, such as switching pumps or opening valves influence sensor readings and downstream behaviour with short but noticeable delays that cannot be ignored.  

Due to the increased digitisation of \ac{IWTS} devices, water treatment systems have also become a target for sophisticated cyber
threats such as logic-layer deception attacks \cite{abraham2025towards,aias20240006,gulzar2024analytical}. Attackers introduce minor system disruptions that preserve numerically plausible device values while altering system behaviour  \cite{rub2024continual,homaei2025causal,urbina2016limiting,hu2019detecting,taormina2017jwpm}.  In such scenarios, \ac{PLC} control logic can be
misled through small, intentionally crafted manipulations, such as altered sensor readings or timing relationships, which remain within acceptable operational limits but gradually drive the process toward unsafe states without triggering immediate alarms \cite{urbina2016limiting,lee2025anomaly}. Similarly, adversaries exploit system dynamics, sensor dependencies, and control logic relationships rather than causing abrupt deviations. As such, logic-layer deception attacks belong to a broader class of stealthy cyber-physical attacks where changes are neither rapid nor easily observable, but which in water treatment facilities can lead to catastrophic consequences affecting public health and environmental safety  \cite{aslam2025gwoae,tuptuk2021systematic, hu2019detecting,mathur2016swat}.

Traditional protection methods against logic-layer deception attacks include threshold-based alarms, rule-based checks, and network intrusion detection systems \cite{hindy2018improving,tuptuk2021systematic, mathur2016swat}.  Similarly, some studies also use model-based methods that depend on physical or control system
models \cite{mathur2016swat,kang2016model,urbina2016limiting}. 
However, prior studies demonstrate that the stealthy attacks described above can evade conventional detection by maintaining statistically normal observations while misleading correct control decisions \cite{amin2013tcst,urbina2016limiting,taormina2017jwpm}. This makes detection that is  based purely on thresholds or statistical deviations insufficient~\cite{lee2025anomaly}. 

Comparatively, anomaly detection approaches based on \ac{ML} and \ac{DL} have been widely used
as a security mechanism to identify suspicious or inconsistent process behaviour from time-series sensor
data, as they can capture patterns that static rule-based methods fail to detect \cite{habib2026cybersecurity,abraham2025towards, aias20240006,tuptuk2021systematic}. Our analysis of existing literature (Section~\ref{sec:1}) shows that many \ac{ML}-based methods are
designed for servers or powerful machines to  detect logic-layer deception attacks, however, these methods are not suitable for \ac{PLC}-class
devices which are inherently resource-constrained with limited memory and processing
power \cite{inoue2017anomaly,aias20240006,wang2023survey}.  Moreover, most existing approaches focus on detecting statistical deviations or global system anomalies, with limited attention to violations in local control logic implemented within \ac{PLC}s that govern sensors and actuators relationships, particularly under resource-constrained environments \cite{lee2025anomaly,aias20240006}. Protecting \ac{PLC}s
from logic layer deception attack is challenging and has not received significant
research attention.
These limitations highlight the need for a light-weight, anomaly detection approach that can operate effectively within \ac{PLC}-based \ac{IWTS} to detect logic-layer anomalies.

To address this gap, this research explores the potential of \acf{TinyDL} to provide light-weight, on-device security awareness for anomaly detection \cite{somvanshi2025tiny}. Instead of using large and
complex models, \ac{TinyDL} focuses on simple and efficient designs that can work
within the strict limits of embedded industrial
devices that require very little memory, computing power, and
energy \cite{somvanshi2025tiny}. \ac{IWTS} are better understood as a sequence of states over time, rather than isolated readings \cite{inoue2017anomaly}, we adopt \textit{\acf{LSTM}} as the core detection
model in this work. \ac{LSTM} is a type of recurrent neural network
designed to work with sequence data, and provides a balance between temporal learning
capability and model compactness   \cite{hochreiter1997long}. It is a stable and compact model, and works well
with short time windows, making it suitable for \ac{TinyDL}-based designs and
deployment on \ac{PLC}-class devices.  

Based on these considerations, the aim of this study is to propose a novel \acf{Ti-iLSTM} framework that uses short time sequences of sensor and
actuator data to detect logic-layer anomalies in \ac{PLC}-based \ac{IWTS}. To achieve this, we address the following research questions:

\begin{itemize}
	\item RQ1: Which \ac{ML}/\ac{DL}-based anomaly
	detection approaches for \ac{IWTS} are
	reported in the literature, and what limitations do they exhibit when deployed
	on resource-constrained \ac{PLC}-class devices?
	\item RQ2: How can a \ac{ML}/\ac{DL}-based anomaly detection model be designed
	and optimised to identify logic-layer deception attacks within \ac{PLC}-based  \ac{IWTS} devices?
	\item RQ3: How can the proposed \ac{TinyDL}-based detection model, in RQ2, be evaluated for its feasibility and  effectiveness with respect to the works identified in RQ1?
\end{itemize}

\begin{figure}
	\centering
	\includegraphics[width=\linewidth]{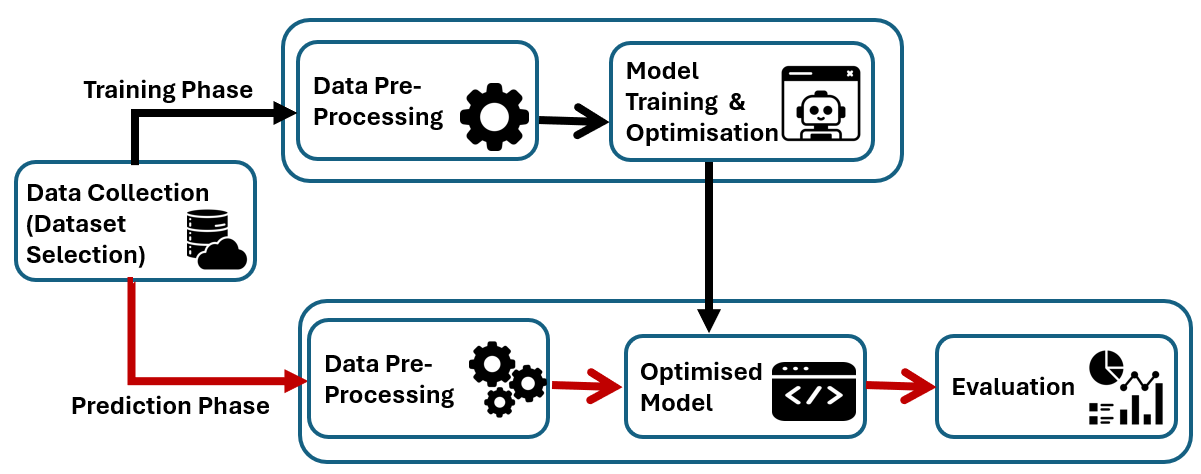}
	%
\caption{Overview of the proposed Ti-iLSTM framework}
\label{fig:framework}
\end{figure} 

For our study we define anomalies as logic-level inconsistencies in system behaviour, where expected cause-and-effect relationships among sensors, actuators, and control logic are violated. Such inconsistencies may arise due to stealthy attacks, sensor manipulation, actuator faults, or process disturbances \cite{homaei2025causal,urbina2016limiting,taormina2017jwpm}; moreover, this  perspective emphasises interaction among components over time rather than solely relying on numerical deviations that focus on individual signal deviations \cite{urbina2016limiting}. 

Fig.~\ref{fig:framework} shows an overview of the proposed \ac{Ti-iLSTM} framework. This framework follows a typical \ac{ML} workflow, and consists of three primary phases: data collection, training (logic-layer anomaly
detection training), and prediction (logic-layer anomaly detection evaluation). \ac{Ti-iLSTM} utilises an optimised incremental \ac{LSTM} model that has enhanced efficiency, low
resource usage,  and smaller size  when compared to 
traditional \ac{LSTM} models for logic-layer detection in \ac{PLC}-based \ac{IWTS}. In this study, \ac{TinyDL} is implemented through resource-aware design choices, including optimal feature selection, short temporal windows, incremental chunk-based processing, and a light-weight model configuration (see Sections~\ref{framework} and~\ref{sec:setup}).

The model is trained and optimised  using the SWaT
dataset \cite{mathur2016swat}, which represents a real industrial water
treatment system. The optimised \ac{TinyDL} model is validated on the WADI dataset \cite{ahmed2017wadi} to
demonstrate that the proposed framework generalises to a different industrial
environment. Please note: the terms \ac{TinyDL} and optimised model can be used interchangeably in this study.

The main contributions of this paper are:
\begin{enumerate}
\item An evaluation of existing \ac{ML}/\ac{DL} anomaly detection mechanisms in the context of \ac{IWTS}
(Section~\ref{sec:1}).
\item Creation of logic-derived anomaly labels using simple  process rules to determine the violation of expected relationship between sensors and actuators, without relying on dataset-provided attack annotations 
(Section~\ref{framework}).
\item A compact three-stage feature selection pipeline that reduces the input
space to an optimal process-relevant variables (Section~\ref{framework}).
\item A light-weight incremental \ac{LSTM} trained using short sliding windows and
chunk-based updates to keep memory usage bounded (Section~\ref{framework}).
\item Experimental validation on a physical \ac{PLC}-class hardware platform with validation time
and \ac{RSS} memory measurement during inference/prediction (Section~\ref{sec:setup}).
\end{enumerate}


\section{Related Works}\label{sec:1}

\begin{table*}[!th]
\caption{Comparative analysis of existing ML/DL-based anomaly detection approaches for \ac{IWTS} (Feat.= Number of features where NR= Not Reported, Win.= Window size, Acc.= Accuracy, F1= F1-score, PLC-D= Solution deployed on \ac{PLC}).}
\label{tab:litreview}
\scriptsize
\begin{tabular}{|p{1cm}|p{1.3cm}|p{3cm}|p{0.7cm}|p{0.7cm}|p{1.2cm}|p{1.5cm}|p{3cm}|p{0.8cm}|} \hline
	\textbf{Study} & \textbf{Dataset} & \textbf{Model} &
	\textbf{Feat.} & \textbf{Win.} & \textbf{HW} &
	\textbf{Performance} & \textbf{Detection Focus} & \textbf{PLC-D} \\ \hline
	\cite{bahadoripour2023deep} & SWaT  &  multi-modal DNN  &  NR  & NR & NR &  F1$\approx$0.98
	& Multi-modal attacks &No  \\ \hline
	
	\cite{boateng2022anomaly} & SWaT &  One-Class SVM    & 51
	& NR & GPU &  F1 $\approx$0.812
	& Unsupervised anomaly detection of process deviations & No  \\ \hline
	\cite{inoue2017anomaly}    & SWaT      & DNN + One-Class SVM
	& 52 & 100 & Server & F1$\approx$0.80
	& Temporal process behaviour & No  \\ \hline
	
	\cite{aias20240006} & SWaT & LSTM + RF + SVM + KNN
	& 51 & 20 & Server & Acc.$\approx$0.98, F1$\approx$0.97
	& Cyberattack classification & No \\ \hline 
	\cite{homaei2025causal} & SWaT/ WADI      & Digital Twin + ML & NR
	&  NR & GPU &  F1$\approx$0.94
	& Anomalies in system behaviour & No  \\ \hline
	\cite{macas2019unsupervised}& SWaT     & STAE-AD (AE+Attn +ConvLSTM)
	& 51 & 120s & Server & F1$\approx$0.88
	& Cross-sensor correlations & No \\  \hline
	\cite{hindy2018improving}       & SCADA     & ML Ensemble (SIEM)
	& NR & $\sim$1s & Server & Acc.$\approx$0.94\%
	& Abnormal SCADA event detection & No \\ \hline
	\cite{liu2025research}      & SWaT/ WADI & VAE--LSTM
	& 51/123 & NR & GPU & F1$\approx$0.75
	& Latent encoding + temporal & No \\ \hline
	\cite{raman2024fusing}     & SWaT      & Engineering inv.+ML
	& NR & NR & NR & NR
	& Process rules + behaviour & No \\ \hline
	\cite{jo2024edge}          & SWaT/ WADI & Graph NN (edge-cond.)
	& 51/127 & 5 & GPU & F1$\approx$0.85
	& Sensor/actuator relations & No \\ \hline
	\cite{xu2023masked}        & SWaT/ WADI & Masked GNN
	& 51/123 & 5 & GPU & F1$\approx$0.83
	& Correlation by masking & No \\ \hline
	\cite{aslam2025gwoae}  & SWaT/ WADI & GWO + Autoencoder
	& NR & NR & Server & Acc.$\approx$0.99
	& Feature sel.+reconstruction & No \\ \hline
	\cite{aly2025shap}         & Industrial & Transformer+GWO+ SHAP
	& NR & NR & Server & Acc.$\approx$0.98, AUC.$\approx$0.996
	& \ac{IIoT} detection +  explainability & No \\ \hline
	\cite{aly2025logboost}     & Industrial & Logistic Boosting
	& NR & NR & Server & Acc.$\approx$0.96, F1.$\approx$ 0.941
	& Classical ML anomaly & No \\ \hline
	\cite{yang2025cloud}       & WADI      & Cloud-edge GNN
	& NR & NR & Edge+ Cloud & NR
	& Edge/cloud collaborative & No \\ \hline
	\cite{s24186131}      & SWaT      & Light-weight 1D-CNN
	& NR & NR & Embedded & Acc.$\approx$0.96  & light-weight embedded & NR \\ \hline
	\textit{This work }        & \textit{SWaT/ WADI} & \textit{Ti-iLSTM}
	& 10 & 10 & \textit{PLC-HW}
	& \textit{F1$\approx$0.96, Acc.$\approx$0.95}
	& \textit{Logic-layer inconsistencies} & \textit{Yes} \\ \hline
\end{tabular}
\end{table*}


This section discusses the existing anomaly detection 
approaches for \ac{IWTS}, with a focus on
how \ac{ML}, including \ac{DL}, and hybrid (combination of \ac{ML} and \ac{DL}) techniques have been applied
to detect cyber-physical and logic-layer anomalies under industrial deployment
constraints \cite{tuptuk2021systematic}. 

Relevant literature was collected from digital libraries: IEEE Xplore, Springer, and ScienceDirect with the 
focus on studies published between 2016 to present to ensure recent and relevant coverage. Table~\ref{tab:litreview} presents a
comparative analysis of representative approaches reported in
the literature. Anomaly detection studies based solely on statistical approaches, Federated learning, Large Language Models (LLMs), and Agentic AI are out of scope.
%



Our literature review (Table~\ref{tab:litreview}) has shown that most of the existing studies used both
supervised and unsupervised \ac{ML}/\ac{DL} algorithms to detect anomalies in water treatment systems by
learning patterns from historical process
data, with a focus on the publicly
available SWaT and WADI benchmark datasets.
A study by Inoue et al. \cite{inoue2017anomaly} focused on sequence-based modelling on the SWaT dataset and showed that using temporal context improves anomaly detection compared to point-based methods. Building on this idea, Komol et al. \cite{aias20240006} applied LSTM-based models along with traditional classifiers and achieved high detection accuracy on SWaT. However, their approaches used all available features in the dataset and were evaluated on server-based systems, without considering resource constraints of \ac{PLC}-class devices. 
Similarly, Muzamil et al. \cite{aslam2025gwoae}  combined optimised
feature selection with reconstruction-based learning, reporting strong
detection performance on SWaT and WADI, but these
evaluations assume server-class execution and do not quantify \ac{PLC}
feasibility. Classical \ac{ML} baselines
remain relevant due to simpler structures, although most studies evaluate them
outside \ac{IWTS} contexts and without deployment
constraints \cite{aly2025logboost}. Overall, most \ac{ML}/\ac{DL}-based detection models
still rely on large feature sets, long windows, or centralised computation,
which reduces suitability for \ac{PLC}-class
deployment.


To better understand complex spatial and temporal
behaviour in water treatment systems, \ac{DL} models such
as autoencoders, \ac{LSTM}-based networks, and attention-based methods are the focus of a number of research articles. These models can learn
detailed patterns from multivariate time-series data and capture the changes in system
behaviour over time. 

Macas et al. \cite{macas2019unsupervised} proposed unsupervised spatio-temp-oral autoencoder model (STAE-AD) that combines convolutional and LSTM layers to capture both spatial and temporal dependencies. The model achieved an F1-score of approximately 0.88 on the SWaT dataset using 51 features and a 120-second window. Likewise, Liu et al. \cite{liu2025research} proposed a variational autoencoder (VAE-\ac{LSTM}) fusion model for wastewater treatment anomaly detection, combining latent feature learning with temporal sequence modelling. Despite the strong detection
capabilities, deep learning and spatio-temporal models introduce substantial
computational and memory overhead, with training and inference typically
performed on GPU-enabled servers and models often requiring large input windows
and extensive parameter sets. Similarly, graph-based deep learning methods  model sensors and actuators as nodes
connected through physical or functional relationships \cite{jo2024edge,xu2023masked}. While these approaches
report strong detection results on SWaT and WADI, they require heavy matrix
operations, message passing, and high-dimensional feature representations,
which makes them less suitable for deployment on resource-limited \ac{PLC}-class
hardware.

Furthermore, several studies have explored anomaly detection in related 
industrial contexts. For example, Hanan et al. \cite{hindy2018improving} applied an \ac{ML} ensemble 
within a Security Information and Event Management (SIEM)  pipeline for SCADA-based water infrastructure, achieving 
approximately 94\% accuracy. Raman et al. \cite{raman2024fusing} proposed a fusion of engineering process invariants with learned behaviour for SWaT anomaly detection, where physical system rules are combined with data-driven learning. However, this multi-module design increases computational complexity and memory usage, which makes it difficult to deploy on PLC or edge devices that have limited resources. Similarly, Aly et al. \cite{aly2025shap} explored a transformer-based model with Grey Wolf Optimiser (GWO) boosting and SHAP (SHapley Additive exPlanations), which is a method used to explain model predictions. The model reports high detection performance, but it is designed for server-class hardware with high processing power and memory. Such architectures are not suitable for \ac{PLC}-based or edge deployment, where models need to be small, fast, and resource-efficient. Also, Yang et al. \cite{yang2025cloud} proposed a collaborative cloud-edge 
gated graph neural network for water distribution anomaly detection, 
reporting improved detection over baselines, though the approach depends on 
cloud partitioning and is not deployable on a standalone \ac{PLC} device. 
Bahadoripour et al. \cite{bahadoripour2023deep} used a deep multimodal cyber-attack detection model focused on network
and sensor modality data from ICS. Homaei et al. \cite{homaei2025causal} identified anomalies in system behavior through causal dependencies by emphasis on interpretability. 

Only a few studies have explored light-weight anomaly detection
specifically designed for embedded or edge industrial
environments. For example, Liu et al. \cite{s24186131} proposed a light-weight CNN model and achieved high accuracy using SWAT dataset. Likewise, in  \cite{antonini2023adaptable}, 
adaptable unsupervised TinyML anomaly detection system was proposed for extreme
industrial conditions, demonstrating that compact models can work under tight
resource constraints. However, while TinyML models are easy to design they are ineffective in dealing with
diverse random attack traffic profiles. 
Zahid et al. \cite{zahid2025light} proposed a light-weight slow-rate attack detection framework for industrial cyber-physical system that measures runtime feasibility along with detection performance. Our work is inspired by their study, which highlights the importance of optimising models to achieve a balance between accuracy and efficiency in resource-constrained \ac{PLC} environments.

Overall, existing anomaly detection approaches for \ac{IWTS}
based on \ac{ML} and DL models prioritise
detection accuracy solely while relying on large feature sets, long temporal windows,
and server-class execution
assumptions. In addition to this, most
studies do not report \ac{PLC}-relevant feasibility evidence such as CPU usage,
memory footprint, or on-device stability. The
approach proposed in this paper addresses these gaps by introducing a novel
light-weight \ac{TinyDL} framework based on an incremental \ac{LSTM} that
operates on fewer optimal features and short temporal windows. It 
measures memory footprint and validation time which infers the CPU utilisation during execution and we validate
generalisation of our approach using publicly available
datasets.


\section{Proposed Framework} \label{framework}

This section presents the \ac{Ti-iLSTM} framework, a light-weight \ac{TinyDL}-based
incremental \ac{LSTM} framework designed to detect logic-layer anomalies in \ac{PLC}-based \ac{IWTS}. The framework
consists of three sequential phases: data collection (dataset selection), training, and prediction, shown in Fig.~\ref{fig:framework}. Each of these phases are presented in the following sections.

%

\subsection{System Representation and Dataset Selection}
\label{subsec:sysrep}

The \ac{IWTS} is represented as a discrete-time process observed through logged
sensor and actuator data. Each system state at timestamp $t$ is captured as a
feature vector ($\mathbf{x_{t}}$) where $x_{t,i} \in \mathbf{x_{t}}$  represents the values at time $t$ of feature $i$. These features correspond to a sensor measurement or actuator state.
\begin{equation}
\mathbf{x_{t}} = [x_{t,1},\, x_{t,2},\, \ldots,\, x_{t,n}] 
\label{eq:feature_vector}
\end{equation}
 Under normal operation,
the temporal evolution of the system state reflects control semantics such as
tank level changes following pump activation, or flow variations following
valve transitions \cite{urbina2016limiting}. Anomalies may still occur even
when individual values remain numerically plausible, if the joint behaviour
violates expected operational
dependencies \cite{inoue2017anomaly}.

The primary dataset used in this work for training the model is the Secure Water Treatment (SWaT)
testbed (July~2019 release (v2)) \cite{mathur2016swat}. 
It is a publicly available benchmark that captures
time-series measurements, logged at one-second intervals, from a fully functional industrial water treatment plant controlled by multiple \ac{PLC}s. The 
SWaT dataset contains 51 sensor and actuator variables recorded 
simultaneously across six stages of a water-treatment testbed. The dataset includes sensor readings such as water levels (LIT), flow rates (FIT), differential pressure (DPIT), and chemical analysers (AIT), along with actuator states for pumps and motorised valves. The Water Distribution (WADI)
dataset \cite{ahmed2017wadi} is used separately for deployment-style validation without retraining.



\subsection{Data Pre-Processing}

\begin{figure}
\includegraphics[width=\columnwidth]{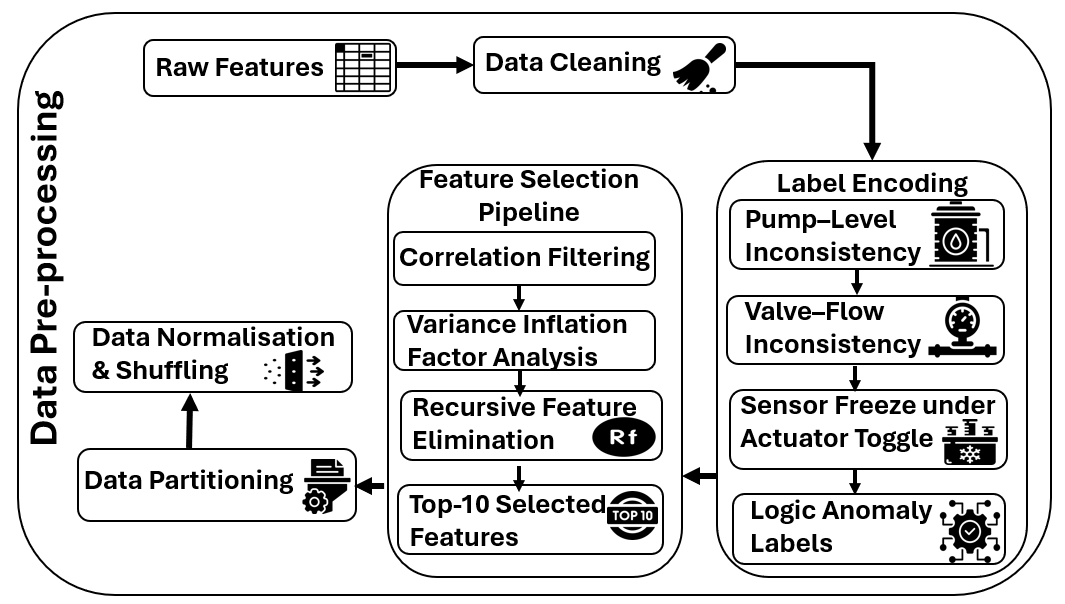}
\caption{Data pre-processing during training phase}
\label{fig:feature_pipeline}
\end{figure}

Data pre-processing comprises data cleaning, feature
selection, data partitioning, data normalisation and shuffling.   Fig.~\ref{fig:feature_pipeline} shows the data pre-processing steps performed during this research. 

\subsubsection{Data Cleaning}
The first step in data pre-processing is to analyse and improve the
raw data quality before proceeding with modeling, analysis, or any related methods. During this step, we performed an in-depth examination to gather statistical information regarding the data’s properties, types, missing input and infinite values, null values, and any columns containing zeros.

\subsubsection{Label Encoding}
A key contribution of this work is that anomaly labels are constructed
without relying on dataset-provided attack annotations. The labels in this study are
derived from interpretable process rules that capture violations of expected
cause-and-effect relationships between sensors and
actuators \cite{urbina2016limiting}. This approach grounds the detection
objective directly in operational process semantics, making the framework more
generalisable to real environments where explicit attack labels are rarely
available. We define three logic-layer inconsistency patterns used for anomaly label construction: pump-level mismatch, valve-flow mismatch, and sensor
freeze during actuator state
transition, as follows: 



\textit{1.Pump-level inconsistency:} When pump (P101) is in the ON state, the
connected tank level sensor (LIT101) is expected to show a measurable change
within a short observation window. If the level remains static beyond a small
tolerance threshold, this indicates a violation of the expected pump-level
relationship and is flagged as anomalous.

\textit{2. Valve-flow inconsistency:} When the motor's valve (MV101) reports an
open state, the downstream flow sensor (FIT101) should register a non-zero flow
reading. If FIT101 remains near zero while MV101 indicates open, this
mismatch represents a breakdown in expected valve-flow behaviour.

\textit{3. Sensor freeze during actuator transition:} If an actuator changes its
operational state, for example pump P101 switching between ON and OFF, the
associated sensor such as LIT101 is expected to respond accordingly. Sensor
readings that show little or no change (within a small tolerance margin) across an actuator state transition are treated as freeze behaviour and flagged as anomalous.

Tolerance margins (details in Section~\ref{sec:setup}) are applied in all three rules to prevent normal sensor
noise and short stabilisation delays from generating false labels. At each
timestamp $t$, each inconsistency condition is evaluated independently and
converted into a binary indicator. The final logic-anomaly label $y_t$ is
computed using a logical OR aggregation:
\begin{equation}
y_t = \max\!\left(\delta^{\text{pump}}_t,\;
\delta^{\text{valve}}_t,\;
\delta^{\text{freeze}}_t\right)
\label{eq:label}
\end{equation}

\noindent where each $\delta$ term takes value 1 if the corresponding
inconsistency condition is satisfied and 0 otherwise. This conservative
design ensures that any single violation of a fundamental operational
dependency is sufficient to mark the system state as anomalous, prioritising
safety by avoiding suppression of subtle logic-layer
deviations.

\subsubsection{Feature Selection Pipeline:}
\label{subsec:features}
Feature selection is an important step before training any 
model, as it improves its decision making accuracy, and effectiveness while reducing complexity \cite{jaw2021feature}. It focuses on identifying a dataset’s most valuable, significant, and appropriate features by removing irrelevant features, ensuring that only the most important variables are used.  Using all variables in a dataset 
directly as model inputs would increase memory usage, training 
time, and inference cost, which is not suitable for resource-constrained devices with limited processing capacity \cite{
wang2023survey}. In addition to the resource constraints, many of these 
variables carry overlapping information because sensors placed 
in the same process stage often respond to the same physical 
changes. Including such redundant variables does not improve 
detection and can make the model less stable during training 
\cite{guyon2002gene, kim2019multicollinearity}. For these 
reasons, a three-stage feature selection pipeline (Fig.~\ref{fig:feature_pipeline}) was applied 
before training the model to reduce the input space to a compact set of optimal process-relevant features.



\textit{Stage~1: Correlation-based filtering:} The Pearson 
correlation coefficient measures the linear relationship 
between two variables, ranging from $-1$ to $+1$, where 
values close to $\pm1$ indicate strong linear dependence 
\cite{benesty2009pearson}.
The less  informative member of each correlated pair was removed,  reducing redundancy caused by multiple sensors measuring  similar physical quantities in the same process stage.

\textit{Stage~2: \ac{VIF} analysis:} 
\ac{VIF} is a statistical measure that quantifies how much the 
variance of one feature is explained by all other features 
combined \cite{kim2019multicollinearity}. For a feature 
$x_i$, its \ac{VIF} is computed as:
\begin{equation}
\text{VIF}_i = \frac{1}{1 - R^2_i}
\end{equation}
where $R^2_i$ is the coefficient of determination obtained 
by regressing $x_i$ on all other features. A \ac{VIF} value of 1 
means the feature is completely independent. Values above 10 
indicate strong multicollinearity, meaning the feature's 
information is already captured by other variables and it can 
be safely removed without losing detection capability 
\cite{kim2019multicollinearity}. \ac{VIF} scores are computed for 
the remaining features to identify multicollinearity. Features 
with a \ac{VIF} value above the accepted threshold are removed 
iteratively until all remaining features exhibit acceptable 
independence \cite{kim2019multicollinearity}.

\textit{Stage~3: Recursive Feature Elimination (RFE) with Random Forest.}
A Random Forest classifier \cite{breiman2001random} is trained on the
logic-derived labels to obtain feature importance scores. RFE then removes
the least important features one step at a time until the final target number
is reached \cite{guyon2002gene}. The Random Forest is used only as a feature
relevance estimator and is not deployed as the final detection model.

This hybrid nature of feature reduction helps keep the model light-weight while still capturing the main operational behaviour related to logic-layer inconsistencies. This pipeline produces a final compact set of ten features (details in Section~\ref{fs}), covering flow rates, tank levels, valve positions,
chemical analyser readings, differential pressure, and pump state. Together
these variables preserve the critical cause-and-effect relationships required
for logic-layer anomaly detection across different stages of the water
treatment process.

\subsubsection{Data Partitioning} 
The training dataset is split into 70\%:30\% training set, and validation set using stratified
partitioning  to maintain consistent class ratios across
all splits \cite{kohavi1995study}. Class imbalance is addressed after feature selection using the Synthetic
Minority Oversampling Technique (SMOTE) \cite{chawla2002smote}, which
generates synthetic anomaly samples by interpolating between existing minority
class instances in the feature space. Oversampling is applied only to the
training partition to prevent information leakage \cite{kaufman2012leakage}. The optimised (\ac{TinyDL}) model is evaluated using a different testing
dataset with unequal sample distribution during the prediction phase.

\subsubsection{Data Normalisation and Shuffling} 
All selected features are 
normalised using standard scaling, where each feature is 
transformed to have zero mean and unit variance. This ensures that features with 
larger numerical ranges, such as flow rate measurements, 
do not dominate over features with smaller ranges, such 
as binary valve states, during LSTM training. Normalisation 
parameters are fitted on the training data only and then 
applied to all splits to prevent information leakage 
\cite{kaufman2012leakage}. After applying SMOTE to address 
class imbalance, the dataset is shuffled using a fixed 
random seed before training. This is necessary because 
SMOTE appends synthetic samples at the end of the dataset 
in order. Without shuffling, the model would see all 
original samples first and all synthetic samples later, 
which introduces ordering bias during chunk-based 
incremental training \cite{chawla2002smote}.
\subsection{Incremental \ac{LSTM} Model Confirguration (Model Training and Optimisation)}
\label{subsec:lstm}



\ac{LSTM} is selected as the core detection model because
\ac{IWTS} behaviour is inherently temporal \cite{hochreiter1997long}. Actuator
commands such as pump activation or valve opening do not produce instant
sensor responses; instead, effects propagate through the physical process
after a short delay. 
This means that logic-layer
inconsistencies are often only observable across a short sequence of
consecutive states rather than at a single time instant.

The \ac{LSTM} processes sequences of consecutive feature vectors using a sliding
window of fixed length ($w$). Each input sequence ($S_t$) at timestamp ($t$) is defined as:
\begin{equation}
\mathbf{S}_t = [\mathbf{x}_{t-w+1},\, \mathbf{x}_{t-w+2},\,
\ldots,\, \mathbf{x}_t]
\label{eq:window}
\end{equation}
\noindent and the prediction target is the logic-anomaly label $y_t$  at the
final timestamp of the window where $y_t \in {0,1}$. The \ac{LSTM} cell at each step maintains a memory
state $c_t$ controlled by gates: input ($i_t$), forget ($f_t$), and
output ($o_t$) gates, shown by Eqs.~\ref{eq:forget}-\ref{eq:hidden} \cite{gers2000learning}

\begin{align}
f_t &= \sigma(W_f[\mathbf{h}_{t-1}, \mathbf{x}_t] + b_f)
\label{eq:forget} \\
i_t &= \sigma(W_i[\mathbf{h}_{t-1}, \mathbf{x}_t] + b_i)
\label{eq:input} \\
o_t &= \sigma(W_o[\mathbf{h}_{t-1}, \mathbf{x}_t] + b_o)
\label{eq:output} \\
c_t &= f_t \odot c_{t-1} +
i_t \odot \tanh(W_c[\mathbf{h}_{t-1}, \mathbf{x}_t] + b_c)
\label{eq:cell} \\
h_t &= o_t \odot \tanh(c_t)
\label{eq:hidden}
\end{align}

\noindent where $b$ is the bias vector, $\sigma$ is the sigmoid function, $\odot$ denotes
element-wise multiplication, $h_t$ represents the hidden state at timestamp ($t$), representing the short-term memory of the model, $\mathbf{h}_{t-1}$ is the previous hidden state, $c_t$ is the cell state at timestamp ($t$) to maintains long-term dependencies, and $W_f, W_i, W_o, W_c$ are trainable weight
matrices. At each timestamp, the cell receives the current input ($x_t$), the previous
hidden state ($h_{t-1}$), and the previous memory state ($c_{t-1}$). The forget,
input, and output gates regulate how the model retains, updates, and exposes past
information. This mechanism allows the model to preserve useful temporal
dependencies while removing unnecessary historical information. The sigmoid function produces a probability score between 0 and 1 for each sequence. Our classification threshold then converts this probability into a binary decision. While the sigmoid function provides the likelihood of anomaly, the threshold determines the operational boundary for detection, allowing control over the trade-off between recall and precision. Detailed information about the working of \ac{LSTM} can  found in  \cite{hochreiter1997long, gers2000learning} and is out of scope for the work we describe here.

The proposed incremental training strategy is designed to 
reflect the memory and processing constraints of resource-constrained \ac{IWTS}, where loading the entire dataset into memory at once 
is not feasible \cite{rub2024continual, zahid2025light}. 
Rather than training the LSTM on the full dataset in a single 
pass, the pre-processed training data are 
divided into fixed-size chunks. For each 
chunk, overlapping sequences of length are constructed 
using a sliding window, where each sequence 
captures the temporal evolution of the selected features over 
the window and the label is taken from the final timestamp of 
that window \cite{hochreiter1997long} (see the experimental details in Section~\ref{sec:setup}). 

The \ac{LSTM} model is then 
trained for a single epoch on each chunk before moving to the 
next, preserving the model weights across chunks so that 
knowledge accumulated from earlier data is retained throughout 
training 
This means that at any point 
during training, the model holds only one chunk of sequences 
in memory rather than the full dataset. This approach bounds peak
memory usage during training and mimics the continuous data arrival scenario
common in real industrial environments \cite{zahid2025light}. After each chunk update, the process \ac{RSS} 
memory and validation time are recorded before and after the 
training step
providing indicative 
feasibility evidence for edge-style execution 
Between chunks, 
explicit garbage collection is performed to release memory, 
further supporting stable execution under constrained 
conditions. 
Thus, to reflect \ac{PLC}-class resource constraints, the model is trained using an
incremental chunk-based strategy. 
\begin{figure}
\centering
\includegraphics[width=0.8\columnwidth]{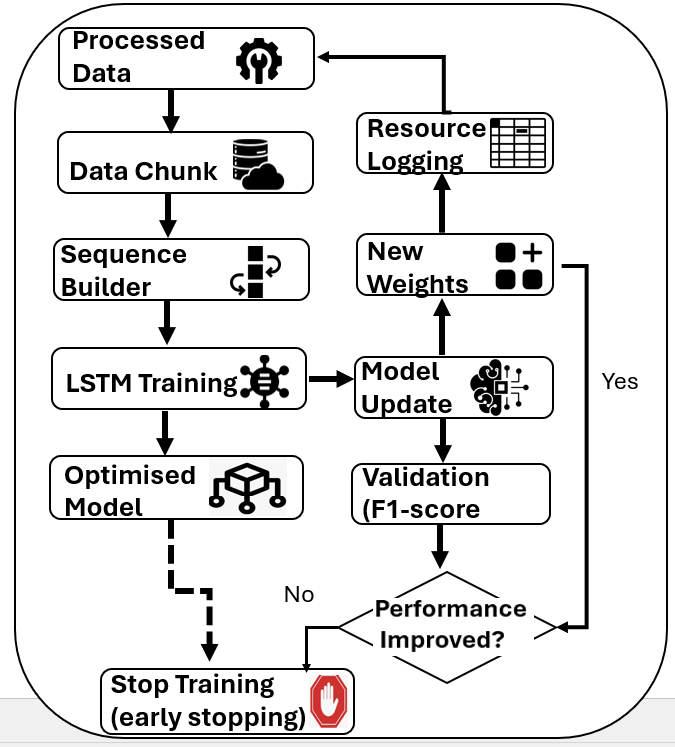}
\caption{Incremental \ac{LSTM} training and optimisation workflow}
\label{fig:ch3_incremental_lstm_workflow}
\end{figure}
The trade-off between model performance and hyper-parameter selection are explored in the results.

Fig.~\ref{fig:ch3_incremental_lstm_workflow} shows the 
incremental training workflow. The workflow consists of 
four steps: sequence builder, \ac{LSTM} training, resource logging, and validation. Each of these steps repeat for each chunk until the final optimised model is obtained. In the first step, the processed (after data pre-processing) data is divided into chunk and each data chunk is passed to the \textit{sequence builder}. The sequence builder constructs overlapping sliding window sequences of 
length $w$ from the normalised feature values as defined in Eq.~\ref{eq:window}. Second, 
the \textit{\ac{LSTM} model} is trained on these sequences, where temporal dependencies are learned through gated operations defined in (Eqs.~\ref{eq:forget}-\ref{eq:hidden}). The forget gate (Eq.~\ref{eq:forget}), input gate (Eq.~\ref{eq:input}), and output gate (Eq.~\ref{eq:output}) regulate the flow of information. Similarly, the cell state update (Eq.~\ref{eq:cell}) integrates long-term and new information, and the hidden state in (Eq.~\ref{eq:hidden}) produces the final output representation. The model is trained for one epoch per chunk, after which a \textit{model update} stores the updated 
weights $W_{i+1}$, which replace the weights $W_i$ 
from the previous chunk. Third, 
\textit{resource logging} records the process \ac{RSS} memory before and after each chunk update as indicated by the resource logging component connected to the model update stage in the Figure~\ref{fig:ch3_incremental_lstm_workflow}. After each update, the workflow loops back to process the next data chunk using the updated (new) weights to ensure the continuous incremental learning across chunks.  Finally, the model is evaluated on the validation set using the F1-score after each chunk update.  
After each chunk update, model performance is evaluated on the validation set using the F1-score. If the score remains same for a fixed number of chunk updates, the training process is stopped. This early stopping criteria prevents unnecessary updates, avoids overfitting to later chunks, and reduce the computational cost \cite{zahid2025light}. This iterative process results in the final optimised \ac{Ti-iLSTM} model, which is then used for inference. 

\subsection{Prediction Phase (Deploying the Framework)}
\label{subsec:evalstrategy}


For deployment-style validation (during the prediction phase), the trained (\ac{TinyDL}) model is applied directly to the
WADI dataset without retraining 
for anomaly detection. The model’s output predicts the anomaly detection task  as a binary classification, where the logic-anomaly label takes value 1 for anomalous behaviour and 0 for normal operation. The same pre-processing pipeline, feature mapping, and
compact feature representation used during SWaT training are applied to WADI
data, ensuring that the validation reflects a realistic generalisation scenario
rather than a dataset-specific re-tuning step. 

Resource feasibility is assessed by recording validation time (inferring the CPU utilisation) and \ac{RSS} memory before and after inference. RSS measures the physical RAM consumed
by the running process and is reported in megabytes. These measurements are
taken during final test inference on both the general-purpose system and the
physical \ac{PLC}, providing indicative evidence of
execution feasibility under bounded resource
conditions. 

\section{Experimental Results and Discussion}
\label{sec:setup}

\begin{figure}
\centering
\includegraphics[width=0.8\columnwidth,height=6cm]{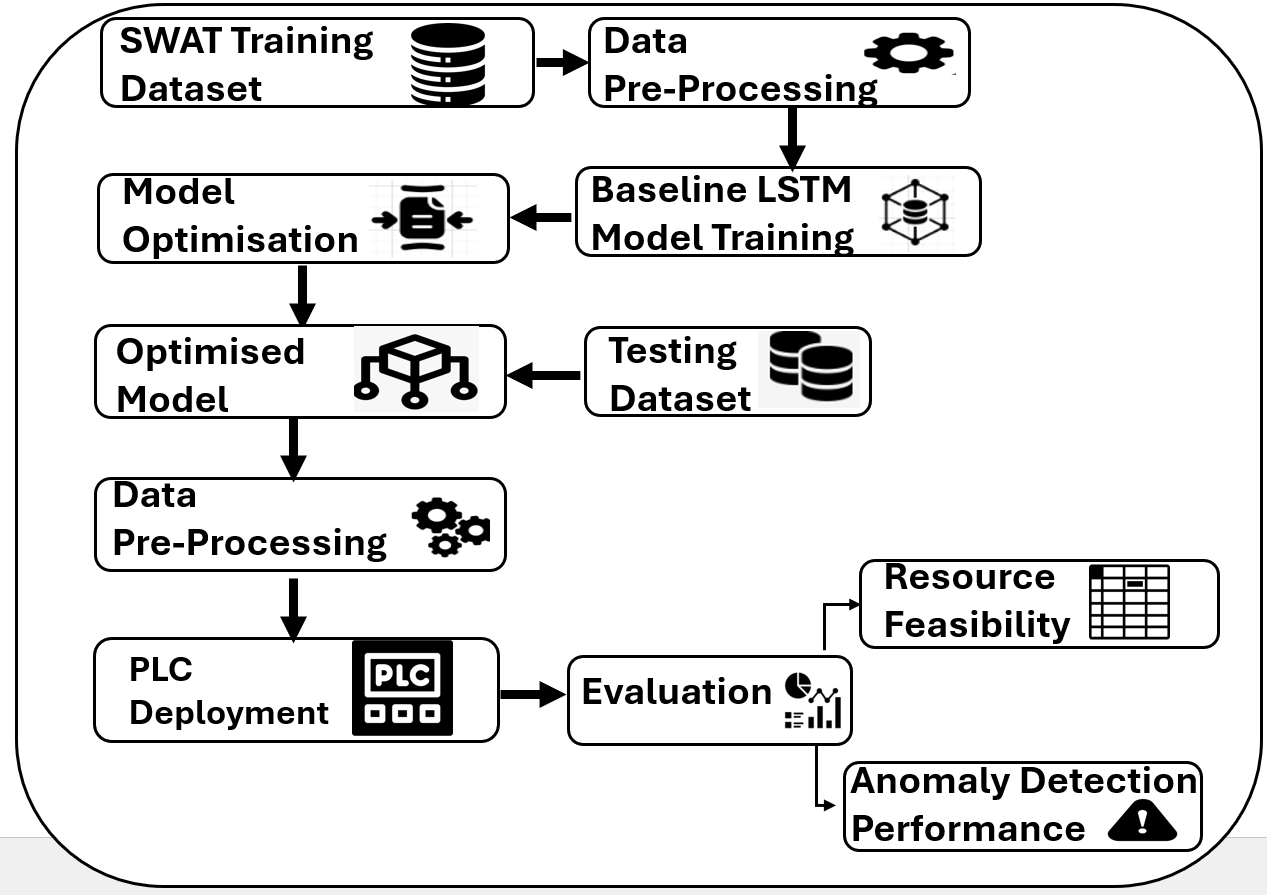}
\caption{Experimental evaluation and prediction workflow for \ac{Ti-iLSTM}}
\label{fig:workflow}
\end{figure}

This section describes the datasets, execution platforms, model
configuration, and evaluation metrics used to assess the proposed \ac{Ti-iLSTM}
framework. It aslo presents the detection performance of the proposed \ac{Ti-iLSTM}
framework on the SWaT dataset, the effect of each light-weight design
parameter, runtime resource usage during inference, cross-dataset validation
on WADI, and a comparison with existing benchmark studies. Fig.~\ref{fig:workflow} shows the experimental training and prediction workflow for the proposed logic-level anomaly detection framework.

The code is available at:  \url{https://github.com/MJ636UoW/TinyML_based_increamental_LSTM}.

\subsection{Execution Platforms}
\label{subsec:platforms}

Two execution environments are used in this study. 
The offline pre-processing, incremental model training, 
and experiments were conducted on an ASUS TUF Gaming A15 
laptop (Windows~11) equipped with an AMD Ryzen~9 5900HX 
processor (3.30\,GHz), 16\,GB RAM, and a 477\,GB SSD. 
The entire implementation was written in Python. 
Specifically, Pandas and NumPy were 
used for data loading and pre-processing, 
scikit-learn for feature selection, stratified 
splitting, and evaluation metrics, statsmodels
for \ac{VIF} analysis and imbalanced-learn for SMOTE-based oversampling. For building and training the incremental 
\ac{LSTM} model, TensorFlow with 
Keras was used. Resource monitoring during inference was 
performed using psutil to record
\ac{RSS} memory consumption \cite{zhang2018pcamp, baciu2024mlino}. 
During the prediction phase, the optimised model was evaluated on a physical \ac{PLC} 
(Industrial Shields Raspberry PLC 50RRA, 4\,GB RAM) to 
determine whether the model can run stably under 
resource-constrained conditions.


\subsection{Analysis of Selected Datasets}
\label{subsec:datasets}

Two publicly available industrial control system benchmarks are used, as
summarised in Table~\ref{tab:datasets}.

\begin{table}[ht]

\caption{Summary of datasets used in experimental evaluation}
\label{tab:datasets}
\renewcommand{\arraystretch}{1.3}
\begin{tabular}{|p{1.5cm}|p{2.3cm}|p{1.5cm}| p{1.6cm}|}
	\hline
	\textbf{Dataset} & \textbf{System Type} & \textbf{Variables} &
	\textbf{Role} \\
	\hline
	SWaT \cite{mathur2016swat} &
	Industrial water treatment plant &
	51 sensor/ actuator tags &
	Training, and validation \\\hline
	WADI \cite{ahmed2017wadi} &
	Industrial water distribution system &
	123 sensor/ actuator tags &
	Deployment validation (inference) without any retraining or feature re-selection \\
	\hline
\end{tabular}
\end{table}


The SWaT dataset is split in the ratio of 70\%:30\% for training and validation purpose, respectively, using a stratified two-stage 
partitioning strategy to maintain consistent class ratios 
across all subsets. 
Stratified splitting ensures 
that the ratio of normal to anomalous sequences is 
preserved across all three partitions.
Table~\ref{binarydataset} shows the number of samples, labels and values of each
label used in the training and testing datasets for binary anomaly detection.
\begin{table}[!h]
\caption{Training and validation dataset information (labels and values, number of samples) for binary logic-layer anomaly detection}
\label{binarydataset}
\scriptsize
	\begin{tabular}{p{0.03\columnwidth}p{0.06\columnwidth}p{0.3\columnwidth}p{0.3\columnwidth}}
		\hline\noalign{\smallskip}
		\textbf{Label} & \textbf{Value} & \textbf{Training (SWaT)~Dataset} & \textbf{Testing (WADI)~Dataset} \\ 
		\noalign{\smallskip}\hline\noalign{\smallskip}
		\multicolumn{1}{l}{\textbf{Benign}} & $0$ & 16,500 & 109,092 \\ 
		\multicolumn{1}{l}{\textbf{Attack}} & $1$ & 13,500 & 162,824 \\ 
		\noalign{\smallskip}\hline
	\end{tabular}
\end{table}

\subsection{Feature Selection} \label{fs}
\begin{figure}
	\centering      
	\includegraphics[width=\columnwidth]{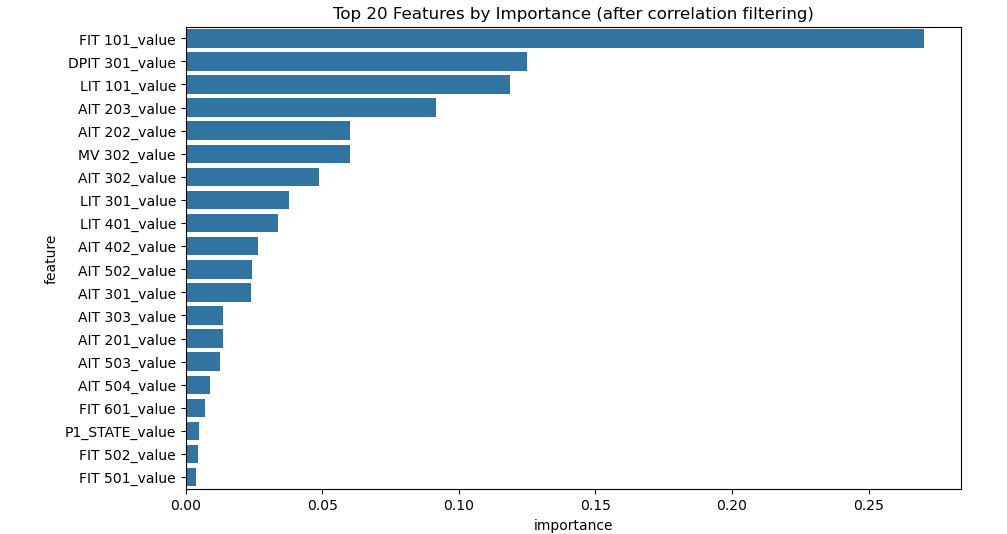}
	\caption{Features selection based on  correlation filtering}
	\label{fig:model2}
\end{figure}

The original SWaT dataset contains 51 sensor and actuator 
variables. In the first stage, a 
pairwise Pearson correlation matrix was computed across all 
51 variables. Any variable whose absolute correlation with 
another exceeded a threshold of 0.9 was removed, as such 
variables carry near-identical information and add no extra 
value to the model. This step reduced the feature pool to 
approximately 20 variables, as shown in Fig.~\ref{fig:model2}. 

In the second stage, \ac{VIF} analysis 
was applied to the remaining variables to check for 
multicollinearity (measures how much the variance of one 
feature is explained by all other features combined). A high 
\ac{VIF} score means the feature is redundant because its 
information is already present in other variables, so it was 
removed to improve model stability and reduce unnecessary 
computation. 
While, in the third stage, a Random Forest classifier with 200 trees 
and balanced class weights was trained on the logic-derived 
anomaly labels to obtain feature importance scores. 
Recursive feature elimination (RFE) 
was then applied to select the optimal ten features for training,removing the least useful variable at each 
step. 
These ten features (Table~\ref{tab:features}) were selected because they cover different 
parts of the water treatment process and together preserve the 
cause-and-effect relationships between actuators and sensors 
that are needed to detect logic-layer anomalies.   

Overall, the chosen features cover different parts of the water treatment process and together they show how the flow changes, how the tank level changes, how the valves work, how the chemicals are measured, and how the pump state changes. This combination preserves
critical cause effect relationships required to identify logic-layer anomalies,
such as pump-level mismatches, valve-flow inconsistencies and abnormal sensor
behaviour during actuator state changes.
\begin{table}[h]
	\caption{Final optimal feature representation}
	\label{tab:features}
	\renewcommand{\arraystretch}{1.2}
	\begin{tabular}{|p{0.5cm}|p{1.37cm}|p{5.2cm}|}
		\hline
		\textbf{ID} & \textbf{Features} & \textbf{Description} \\
		\hline
		F1  & FIT\_101&Influent flow rate measurement \\\hline
		F2  & LIT\_101&Raw water tank level \\\hline
		F3  & MV\_101  & Motorised valve position at intake \\\hline
		F4  & AIT\_202 & Chemical concentration measurement \\\hline
		F5  & AIT\_203 & Chemical analyser in treatment stage \\\hline
		F6  & AIT\_302 & Post-treatment chemical analyser value \\\hline
		F7  & DPIT\_301 & Differential pressure across filtration unit \\\hline
		F8  & LIT\_301 & Intermediate tank level measurement \\\hline
		F9  & MV\_302 & Valve state controlling treated water flow \\\hline
		F10 & P3\_STATE & Pump operational state indicator \\
		\hline
	\end{tabular}
\end{table}

\subsection{Performance Evaluation}
\label{subsec:metrics}

The result of the optimised model (\ac{TinyDL}) is evaluated using detection performance and resource feasibility.  Detection performance includes five standard classification metrics:
accuracy, precision, recall, F1-score, and area under the receiver operating
characteristic curve (ROC-AUC) \cite{fawcett2006introduction, Precisio63:online}:
\begin{equation}
	\text{Accuracy} = \frac{TP + TN}{TP + TN + FP + FN}
	\label{eq:accuracy}
\end{equation}
\begin{equation}
	\text{Precision} = \frac{TP}{TP + FP}, \qquad
	\text{Recall} = \frac{TP}{TP + FN}
	\label{eq:prec_rec}
\end{equation}
\begin{equation}
	\text{F1} = \frac{2 \times \text{Precision} \times \text{Recall}}
	{\text{Precision} + \text{Recall}}
	\label{eq:f1}
\end{equation}
\noindent where $TP$, $TN$, $FP$, and $FN$ denote true positives, true
negatives, false positives, and false negatives
respectively.
In safety-critical
\ac{IWTS} monitoring, recall is especially important because a missed anomaly
carries a higher operational risk than a false
alarm \cite{urbina2016limiting}. 
The F1-score provides a balanced view of both precision and recall, and ROC-AUC measures the model's ability to discriminate between normal and anomalous sequences across all threshold settings.

The resource feasibility is assessed using validation time (s) and \ac{RSS} memory (MB) recorded before and
after final test inference. 
These measurements provide
indicative evidence of execution feasibility under \ac{PLC}-based \ac{IWTS} constraints and
complement the detection performance metrics in a way that is largely absent
from prior literature.
\subsection{Initial Results}
\label{result}

The incremental \ac{LSTM} is trained using the final ten feature representation
described in Section~\ref{fs}.  The baseline configuration used
for all primary experiments is: window length $W = 5$, chunk size $C = 1000$,
batch size $B = 64$, \ac{LSTM} units $U = 16$, and classification threshold
$\tau = 0.5$.







Table~\ref{tab:wadi} shows the results trained on the SWaT dataset and directly
applied to the WADI dataset without any retraining or feature
re-selection. The \ac{TinyDL} model
achieved an accuracy of 0.949 and an F1-score of 0.957, with precision
remaining high at 0.974.

\begin{table}[h]
	
	\small
	\setlength{\tabcolsep}{4pt}
	\caption{Detection performance of \ac{TinyDL} on SWaT and WADI dataset}
	\label{tab:wadi}
	\begin{tabular}{|c|c|c|c|c|}
		\hline
		\textbf{Dataset} & \textbf{Accuracy} & \textbf{Precision} & \textbf{Recall} &
		\textbf{F1-score}  \\
        \hline
		SWaT & 0.98 & 0.974 & 0.989 & 0.981 \\
		\hline
		WADI & 0.949 & 0.974 & 0.94 & 0.957 \\
		\hline
	\end{tabular}
\end{table}

The drop in performance compared to SWaT is expected. WADI is a different
industrial system with a larger number of variables, a different physical
layout, and different process dynamics \cite{ahmed2017wadi}. The model was
not retrained or adjusted for WADI in any way, so this result reflects a true
cross-dataset generalisation
test \cite{tuptuk2021systematic}. The fact that precision
stays high at 0.974 shows that the logic-aware supervision approach captures
process behaviour patterns that remain meaningful across different water
infrastructure settings. This result also shows that, in real-world settings, where delays in anomaly detection can lead to catastrophic consequences, the trade-off between accuracy and performance is significant.

\subsubsection{Resource Usage During Training and  Inference}
\label{subsec:resource}

Table~\ref{tab:resource} shows the memory utilisation across two execution environments: model evaluation on the SWAT test set using laptop environment and inference on the \ac{PLC} using the WADI dataset under the baseline configuration. It has been observed that the memory usage
increased by only 7.65\,MB, from 701.18\,MB to 708.83\,MB
supporting demonstrating that the proposed framework is feasible for \ac{PLC}-class execution.

\begin{table}[h]
	
	\small
	\caption{Comparison of resource usage of optimised model on execution platforms }
	\label{tab:resource}
	\begin{tabular}{|p{2.1cm}|p{2.7cm}|p{2.2cm}|}
		\hline
		\textbf{Metric} & \textbf{Validation (Laptop)} & \textbf{Inference (\ac{PLC})} \\
		\hline
		Process RSS memory (MB) & 701.18 & 708.83 \\\hline
		System memory (\%)      & 81.6   & 81.7   \\\hline
		
	\end{tabular}
\end{table}


\subsection{Model Configuration for Selection of Optimised Model (\ac{TinyDL})}
\label{subsec:config}
    
In the follow results, we explore the impact of parameters on model performance and resource utilisation. In each of the result tables, the highlighted green row indicated the selected final value for that parameter. The goal of this work is to design a light-weight model suitable for \acp{PLC}, configurations that provide good performance with lower resource usage are preferred over configurations that maximize performance at the cost of higher computational overhead. Therefore, The choose value represents a balance between detection performance and resource usage. In several cases, configurations with slightly higher F1-score or accuracy were observed to consume significantly higher memory or require longer execution. This selection strategy ensures that the final model remains practical for deployment rather than only optimised for accuracy.

In this study, eight controlled experiments (E1-E8) were 
designed to evaluate the detection performance and resource 
efficiency of the optimised model on the SWaT 
dataset. E1 is the baseline configuration  
which serves as the 
reference point for all comparisons. E2 and E3 were intentionally 
designed as resource-heavy configurations using larger window 
sizes, batch sizes, chunk sizes, and \ac{LSTM} units to simulate 
non-\ac{PLC}-friendly conditions (characterise the experimental environment, and perform preliminary parameter tuning), therefore, we are not presenting the results for resource-intensive experiments. These two experiments show how 
increased model complexity raises memory usage and validation time, 
which directly justifies the need for the lightweight design 
choices made in this work. 

The remaining experiments E4 through 
E8 each vary one parameter at a time while keeping all others 
at the baseline values. E4 evaluates the effect of window size 
on temporal context and detection performance. E5 studies the 
impact of batch size on training stability. E6 investigates 
how chunk size affects incremental learning and memory behaviour. 
E7 explores the trade-off between model capacity and 
computational cost by varying the number of \ac{LSTM} units. 
E8 examines the precision-recall trade-off by adjusting the 
classification threshold. Together, these experiments isolate 
the effect of each parameter and support the selection of an 
optimal configuration for real-time resource-constrained 
\ac{IWTS}.

\subsubsection{Window Size Determination}

Table~\ref{tab:window} demonstrates how changing the sliding window length impacts the model's performance and resource utilisation during validation. All three configurations maintain high detection performance, with accuracy remaining above 0.983, indicating that the model is not highly sensitive to  changes in window size. However, a clear increase in memory usage is observed as the window size grows, rising from 750.89\,MB at $w=5$ to 862.92\,MB at $w=20$.  Window size $w=5$ achieves better model performance, lower detection time and memory use therefore is the most suitable value.
\begin{table}
	\small
	\setlength{\tabcolsep}{5pt}
	\caption{Effect of window size ($w$) on detection performance and resource usage during validation
		(Chunk=1000, Batch=64, Units=16, Threshold=0.5)}
	\label{tab:window}
    \begin{tabular}{|c|c|c|c|c|c|c|}
    \hline
    \textbf{$W$} & \textbf{Acc.} & \textbf{Prec.} & \textbf{Rec.} & \textbf{F1} & \textbf{RSS}  & \textbf{Total Time} \\
     &  &  &  &  & \textbf{(MB)}  & \textbf{(s)} \\
		\hline
\rowcolor{green!20}		5  & 0.984 & 0.973 & 0.992 & 0.983 & 750.89  & 1.021 \\\hline
10 & 0.983 & 0.972 & 0.991 & 0.981 & 797.80  & 1.217 \\\hline
20 & 0.984 & 0.972 & 0.993 & 0.982 & 862.92 & 1.370 \\\hline
	\end{tabular}
\end{table}

\subsubsection{Batch Size Determination}
Table~\ref{tab:batch} show that as batch size increase so does the cost of increased memory usage. In very large batches such as 128 reduce detection quality, as reflected in the drop in accuracy. Batch size 16 maintaining reliable detection performance without introducing unnecessary memory overhead making it suitable for practical deployment. 


\begin{table}[!h]
\centering
\scriptsize
\setlength{\tabcolsep}{2pt}
\caption{Effect of batch size on detection performance and resource usage during validation
(Window=10, Chunk=1000, Units=16, Threshold=0.5)}
\label{tab:batch}
\resizebox{\columnwidth}{!}{%
\begin{tabular}{|c|c|c|c|c|c|c|}
\hline
\textbf{Batch} & \textbf{Acc.} & \textbf{Prec.} & \textbf{Rec.} & \textbf{F1} & \textbf{RSS}  & \textbf{Total Time} \\
 &  &  &  &  & \textbf{(MB)}  & \textbf{(s)} \\
\hline
\rowcolor{green!20} 16  & 0.987 & 0.974 & 0.998 & 0.986 & 882.51 &  1.198 \\\hline
32  & 0.986 & 0.972 & 0.997 & 0.985 & 916.30 &  1.414 \\\hline
64  & 0.984 & 0.969 & 0.995 & 0.982 & 921.10  & 1.212 \\\hline
128 & 0.970 & 0.975 & 0.957 & 0.966 & 932.09 &  1.204 \\\hline
\end{tabular}%
}
\end{table}


\subsubsection{Chunk Size Determination}

Table~\ref{tab:chunk} shows that detection performance remains consistently high across all chunk sizes, with accuracy staying above 0.982. However, memory usage increases noticeably as chunk size grows, reaching 1019.44\,MB at $C=5000$, which exceeds the practical limits for constrained environments. 

From a runtime perspective, chunk size $C=1000$ maintains low detection time (lower CPU usage), while larger chunks introduce additional processing overhead without significant performance gain. Based on this observation, $C=1000$ is selected as it offers a good balance between detection performance and resource efficiency. 


\begin{table}[!t]
\centering
\scriptsize
\setlength{\tabcolsep}{2pt}
\caption{Effect of chunk size on detection performance and resource usage during validation
(Window=10, Batch=64, Units=16, Threshold=0.5)}
\label{tab:chunk}
\resizebox{\columnwidth}{!}{%
\begin{tabular}{|c|c|c|c|c|c|c|c|}
\hline
\textbf{Chunk} & \textbf{Acc.} & \textbf{Prec.} & \textbf{Rec.} & \textbf{F1} & \textbf{RSS} & \textbf{Total Time} \\
 &  &  &  &  & \textbf{(MB)}  & \textbf{(s)} \\
\hline
500  & 0.984 & 0.974 & 0.991 & 0.982 & 948.30 & 1.366 \\\hline
\rowcolor{green!20}1000 & 0.983 & 0.974 & 0.989 & 0.982 & 986.81 & 1.205 \\\hline
5000 & 0.982 & 0.976 & 0.984 & 0.980 & 1019.44 &  1.185 \\\hline
\end{tabular}%
}
\end{table}

\subsubsection{Final Selection of \ac{LSTM} Units (Model Capacity)}

Table~\ref{tab:units} shows that increasing the number of \ac{LSTM} units from 16 to 64 leads to only a marginal improvement in F1-score (from 0.982 to 0.984), while the impact on memory usage is not significant. This indicates that 16 units are already sufficient to capture the short-term temporal dependencies present in the selected features.

In addition, larger models introduce higher detection time without a meaningful gain in performance. Therefore, selecting 16 units keeps the model compact and efficient, which is important for deployment in resource-constrained environments.

\begin{table}[!t]
\centering
\scriptsize
\setlength{\tabcolsep}{2pt}
\caption{Effect of \ac{LSTM} units on detection performance and resource usage during validation
(Window=10, Chunk=1000, Batch=64, Threshold=0.5)}
\label{tab:units}
\resizebox{\columnwidth}{!}{%
\begin{tabular}{|c|c|c|c|c|c|c|c|}
\hline
\textbf{Units} & \textbf{Acc.} & \textbf{Prec.} & \textbf{Rec.} & \textbf{F1} & \textbf{RSS}  & \textbf{Total Time} \\
 &  &  &  &  & \textbf{(MB)}  & \textbf{(s)} \\
\hline
\rowcolor{green!20}16 & 0.984 & 0.972 & 0.993 & 0.982 & 1053.42 & 1.198 \\\hline
32 & 0.984 & 0.968 & 0.998 & 0.983 & 1031.89  & 1.292 \\\hline
64 & 0.985 & 0.970 & 0.998 & 0.984 & 1041.86  & 1.812 \\\hline
\end{tabular}%
}
\end{table}




	




\subsubsection{Classification Threshold Determination }
The incremental \ac{LSTM} outputs a probability value between 
0 and 1 for each input sequence. This probability represents 
how confident the model is that the sequence contains a 
logic-layer anomaly. A classification threshold is needed to 
convert this continuous probability into a binary decision. 
Any sequence with a predicted probability above the threshold 
is labeled as anomalous, and any sequence below it is labeled 
as normal.
The choice of 
threshold matters because it directly controls the balance 
between detecting anomalies and avoiding false alarms. 
The classification threshold directly affects the trade-off between precision and
recall. In this study, threshold values are selected heuristically to represent
low, balanced, and conservative decision settings, allowing the precision-recall
behaviour to be observed under an identical model configuration.
\begin{table}[!t]
\centering
\scriptsize
\setlength{\tabcolsep}{2pt}
\caption{Effect of threshold on detection performance and resource usage during validation
(Window=10, Chunk=1000, Batch=64, Units=16)}
\label{tab:threshold}
\resizebox{\columnwidth}{!}{%
\begin{tabular}{|c|c|c|c|c|c|c|c|}
\hline
\textbf{Thresh.} & \textbf{Acc.} & \textbf{Prec.} & \textbf{Rec.} & \textbf{F1} & \textbf{RSS} & \textbf{Total Time} \\
 &  &  &  &  & \textbf{(MB)} & \textbf{(s)} \\
\hline
0.3 & 0.980 & 0.959 & 0.998 & 0.978 & 1063.45 &1.205 \\\hline
\rowcolor{green!20}0.5 & 0.98 & 0.974 & 0.989 & 0.981 & 1097.56 &  1.207 \\\hline
0.7 & 0.977 & 0.983 & 0.966 & 0.975 & 1125.45 &  1.197 \\\hline
0.9 & 0.948 & 0.994 & 0.891 & 0.939 & 1355.30  & - \\\hline
\end{tabular}%
}
\end{table}

Threshold values 0.3, 0.5, 0.7, and 0.9 were evaluated to reflect progressively
stricter anomaly decision criteria. Lower thresholds prioritise recall and early
detection, while higher thresholds emphasise precision and reduce false alarms,
which is particularly relevant in safety-critical industrial environments.
Table~\ref{tab:threshold} shows how the decision threshold influences the balance between precision and recall. At a lower threshold of 0.3, recall increases to 0.9988, indicating that most anomalies are successfully detected, but this comes with a drop in precision to 0.9592, leading to more false alarms. On the other hand, a higher threshold of 0.9 improves precision to 0.9940, but recall drops significantly to 0.8910, meaning that several true anomalies are missed.

In the context of a water treatment system, missing an anomaly is more critical than generating additional alerts, as an undetected logic-layer issue can silently affect the physical process and lead to unsafe conditions. Therefore, using a very high threshold is not appropriate.

\begin{figure}
	\centering
	
	\begin{minipage}{0.48\columnwidth}
		\centering
		\includegraphics[width=\linewidth]{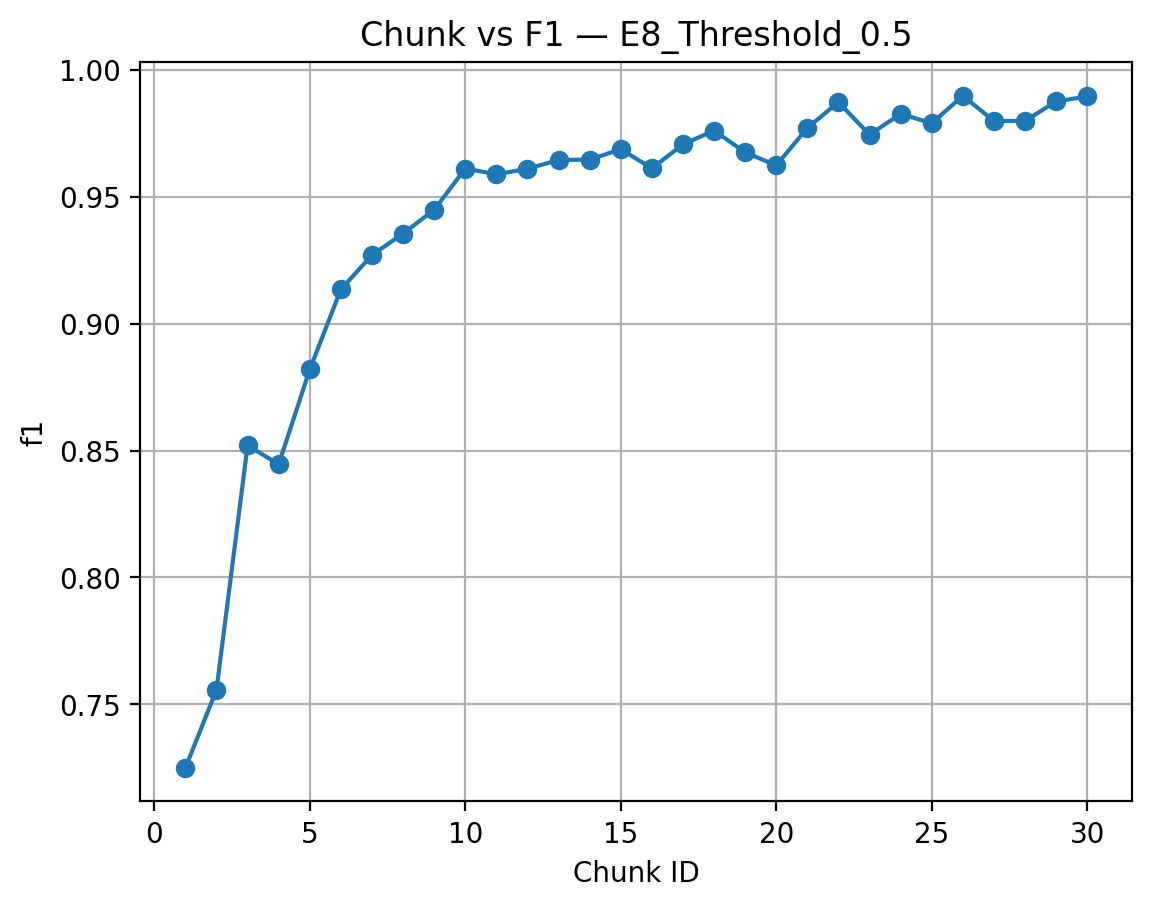}
		
		\small (a) F1 score
	\end{minipage}
	\hfill
	\begin{minipage}{0.48\columnwidth}
		\centering
		\includegraphics[width=\linewidth]{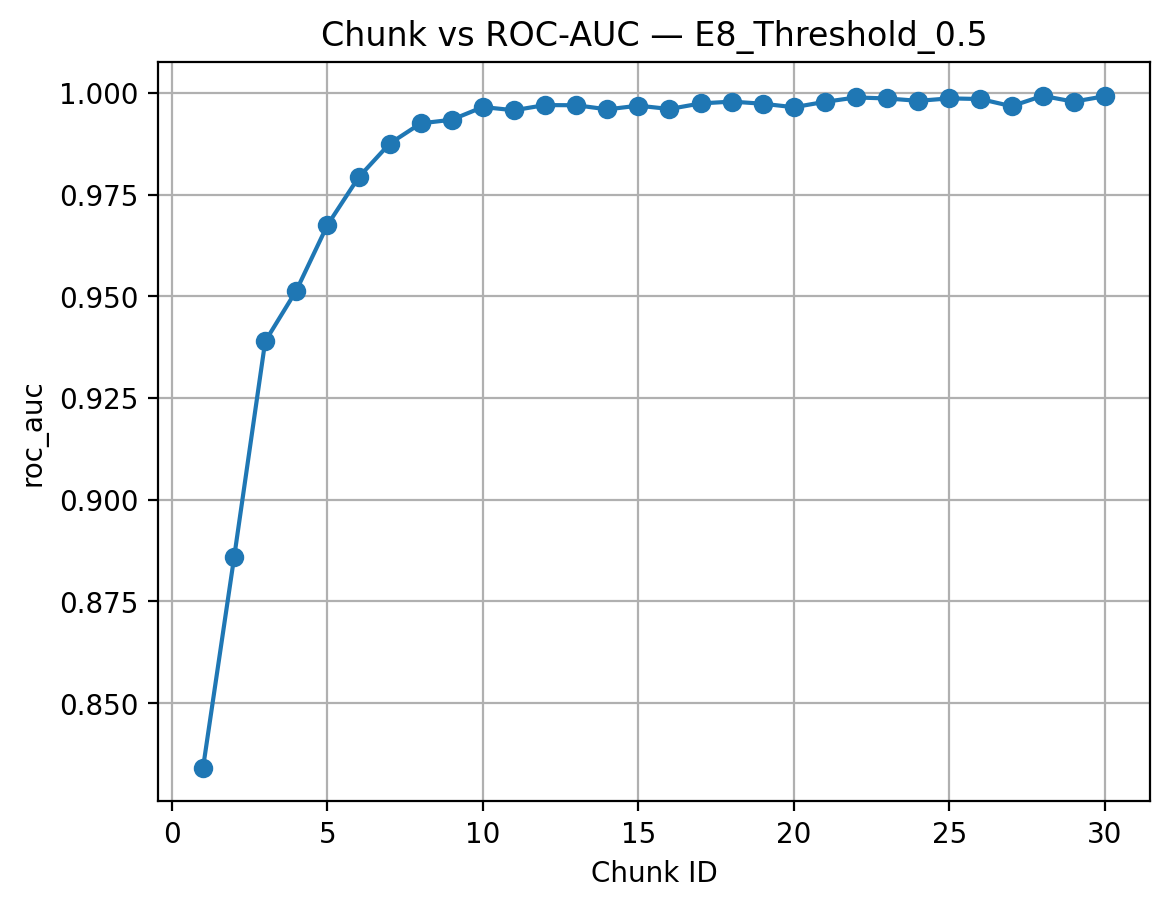}
		
		\small (b) ROC-AUC
	\end{minipage}
	
	\vspace{0.5em}
	
	\begin{minipage}{0.48\columnwidth}
		\centering
		\includegraphics[width=\linewidth]{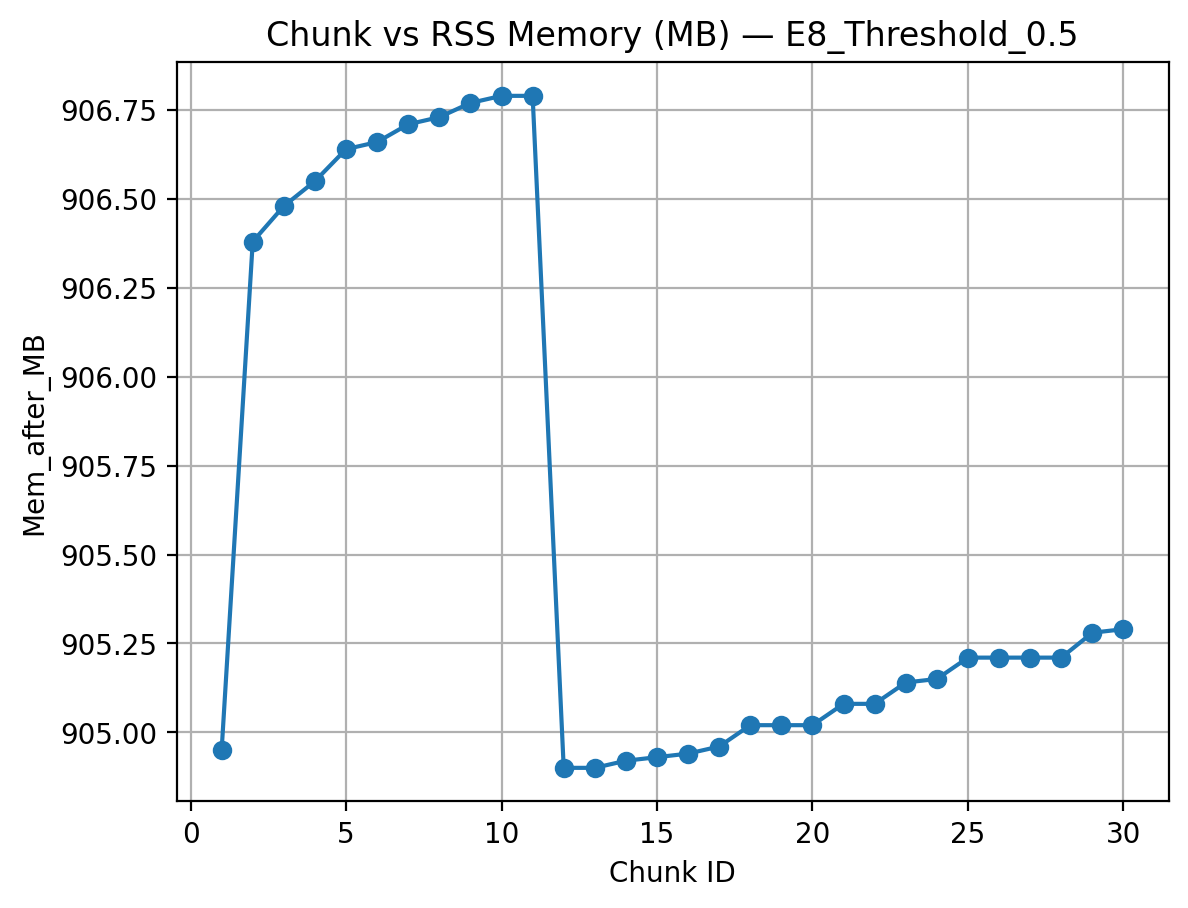}
		
		\small (c) RSS memory (MB)
	\end{minipage}
	\hfill
	%
	
	\caption{E8 threshold experiment (Threshold = 0.5): performance (F1, ROC-AUC) and resource usage (RSS memory) across incremental update chunks on SWaT during training.}
	\label{fig:e8_4plots_threshold05}
	
\end{figure}

A threshold of 0.5 (Fig.~\ref{fig:e8_4plots_threshold05}) is selected as the final setting, as it maintains a good balance between precision and recall while keeping the F1-score consistent and reliable.

\subsection{Baseline Detection Performance on SWAT}
\begin{figure}
	\centering
	\includegraphics[width=\columnwidth]{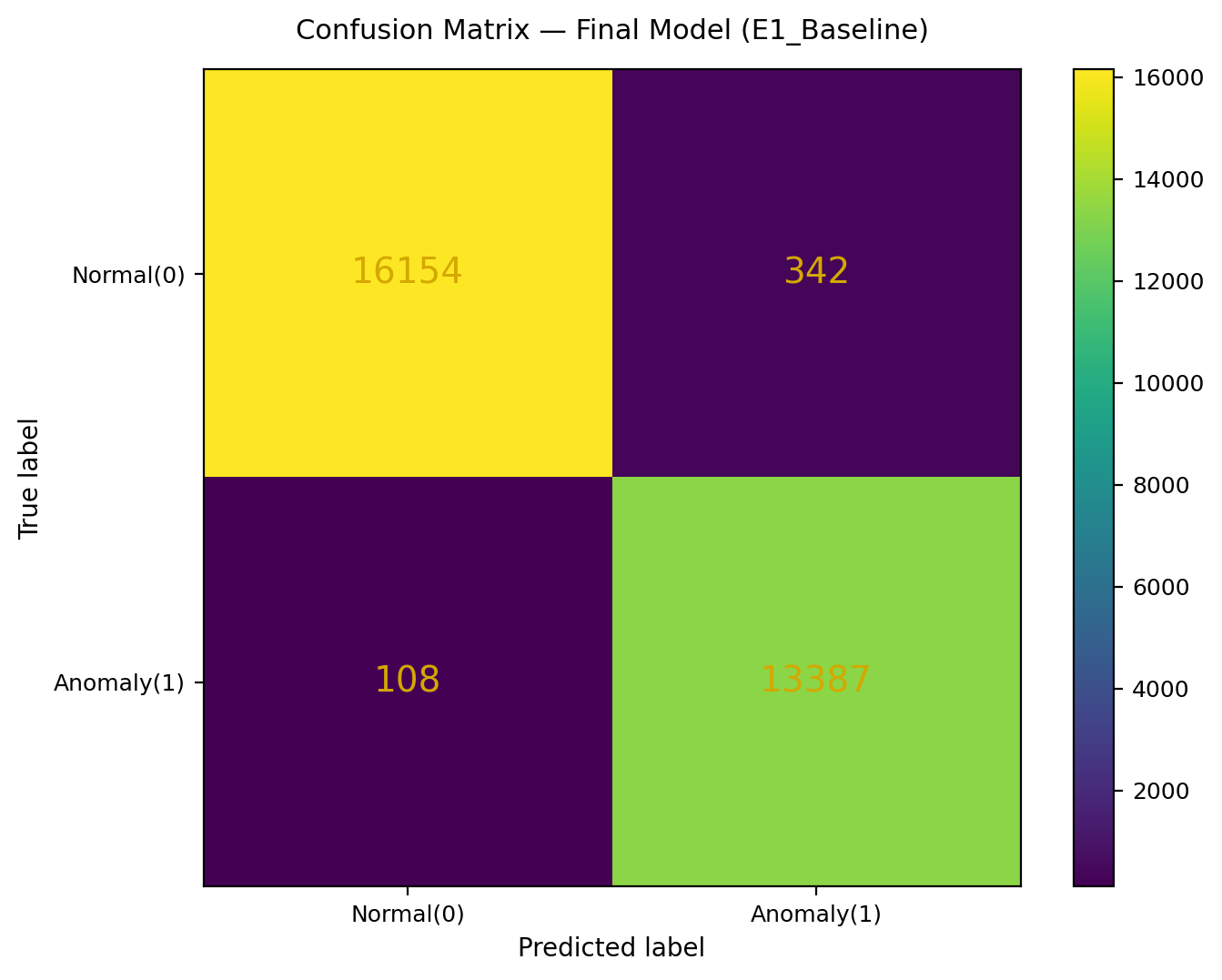}
	\caption{Confusion matrix for the incremental \ac{LSTM} on the SWaT test set}
	\label{fig:confusion_matrix_lstm}
\end{figure}

Using the baseline configuration described in
Section~\ref{subsec:config}, the incremental \ac{LSTM} achieved the results shown
in Fig.~\ref{fig:confusion_matrix_lstm}. 
 The confusion matrix results confirm these findings. The model produced
16,154 true negatives, 342 false positives, 108 false negatives, and 13,387
true positives. Only 108 logic anomalies were
missed out of a total of 13,495 anomalous sequences, showing strong detection
sensitivity. In safety-critical water treatment monitoring, a low false
negative count is more important than a low false positive count, because a
missed anomaly carries a higher operational
risk \cite{urbina2016limiting}. The results confirm that the proposed model
handles this priority well.

\subsection{Comparison with Existing Benchmarks}
\label{subsec:comparison}

Table~\ref{tab:comparison} compares the proposed model against selected SWaT
benchmark studies. Only values that are explicitly reported in each paper are
included. Missing values are marked as NR (Not Reported).

\begin{table}[h]
	\small
	\caption{Comparative analysis against key SWaT benchmarks using explicitly
		reported metrics (NR\,=\,Not Reported, Refer=References, F=Features, Win=Window size, Mem=Memory)}
	\label{tab:comparison}
	\begin{tabular}{|p{0.8cm}|p{1.7cm}|p{0.4cm}|p{0.52cm}| p{0.67cm}|p{1.8cm}|}
		\hline
		\textbf{Refer} & \textbf{Model} & \textbf{F} &
		\textbf{Win} & \textbf{F1} & \textbf{Memory} \\
		\hline
		\cite{macas2019unsupervised} &
		STAE-AD & 51 & 120\,s & 0.88 & NR \\\hline
		\cite{aias20240006} &
		LSTM & 51 & 20 & 0.94 & NR \\\hline
		\cite{aslam2025gwoae} &
		GWO-Autoencoder & NR & NR & NR & NR \\
		\hline
		This work &
		\ac{Ti-iLSTM} (TinyDL) & 10 & 10 & 0.983 &
		RSS$\approx$709\,MB \\\hline
	\end{tabular}
\end{table}

The proposed model achieves a higher F1-score than all benchmark studies that
report this metric. Compared to STAE-AD \cite{macas2019unsupervised} with
F1$\approx$~0.88, the proposed model improves detection while using far fewer
features (10 versus 51) and a much shorter window (10 versus 120 seconds).
Similarly, compared to the \ac{LSTM} 
from \cite{inoue2017anomaly} which achieved F1$\approx$0.94 using all 51 features on server hardware,  our \ac{TinyDL} model, 
\ac{Ti-iLSTM}, presents a clear improvement in detection performance (F1$\approx$0.98) using only ten features. Overall, the key difference
between our work and all benchmark studies is that this work explicitly
reports validation time (inferring CPU usage) and \ac{RSS} memory during prediction. None of the benchmark
papers provide runtime resource measurements, meaning their practical
deployability on \ac{PLC} remains
unknown. This work is therefore the only study in
this comparison that provides both strong detection performance and evidence of
execution feasibility under resource-constrained
conditions.

\subsection{Limitations}
\label{sec:limitations}

While the results are promising, several limitations should be acknowledged: First,  core model training and testing    were performed on logged process datasets rather than on live plant execution,
which limits direct real-time operational claims \cite{mathur2016swat}.     Similarly, the logic-layer anomaly labels
were constructed using interpretable rules that capture key inconsistencies.
However, real plants may include more complex dependencies across stages and
varying time delays, which are not fully modelled in this study. Third, the CPU and memory values were
recorded as snapshots during inference. Full profiling of execution cycles,
hard real-time latency, and direct power measurement were outside the scope of
this work.
\section{Conclusions and Future Directions}
\label{sec:conclusion}

This paper proposed \ac{Ti-iLSTM}, a light-weight \ac{TinyDL}-based incremental \ac{LSTM}
framework for detecting logic-layer anomalies in \ac{PLC}-based \ac{IWTS}. Our work was motivated by a
clear gap in existing research, where most anomaly detection models are
designed for server or GPU environments and do not address the memory and
processing limits of real industrial control
hardware.

The proposed framework makes four contributions. First, anomaly labels are
built from interpretable process rules that capture violations of expected
cause-and-effect behaviour between sensors and
actuators, without relying on dataset-provided
attack annotations. Second, a structured three-stage feature selection
pipeline reduces the input space to ten compact and process-relevant
variables. Third, a
light-weight incremental \ac{LSTM} is trained using short sliding windows and
chunk-based updates to keep memory usage
bounded. Fourth, the model is
validated on a physical \ac{PLC} with explicit measurement of validation time which infers CPU usage and
memory measurement during inference. 
This work is the only
study in the benchmark comparison to report runtime resource feasibility
alongside detection metrics.

Future work will focus on three directions. Model compression through
quantisation and pruning can reduce the memory footprint further for tighter
PLC constraints. The framework can be extended by using Large language models for the explainable analysis of detected logic-level anomaly and generation of human-interpretable security insights. Finally, the logic consistency rules can be expanded to cover more
process stages and more complex cause-and-effect patterns across the full
water treatment workflow.



\printcredits

\bibliographystyle{cas-model2-names}

\bibliography{Tinymain}





\end{document}